# Data Pipeline Training: Integrating AutoML to Optimize the Data Flow of Machine Learning Models

Jiang Wu[1,*]
Computer Science
University of Southern California,
Los Angeles, CA, USA
jiangwu@usc.edu

Hongbo Wang[1]
Computer Science
University of Southern California
Los Angeles, CA
hongbowa@usc.edu

Chunhe Ni[2]
Computer Science
University of Texas at Dallas,
Richardson, TX, USA
nichunhe@outlook.com

Chenwei Zhang[3]
Electrical and Computer Engineering
University of Illinois Urbana-Champaign
Urbana, IL
zchenwei66@gmail.com

Wenran Lu[4]
Electrical Engineering
University of Texas at Austin
Austin, TX
wenranlu@gmail.com

*Abstract*—Data Pipeline plays an indispensable role in tasks such as modeling machine learning and developing data products. With the increasing diversification and complexity of Data sources, as well as the rapid growth of data volumes, building an efficient Data Pipeline has become crucial for improving work efficiency and solving complex problems. This paper focuses on exploring how to optimize data flow through automated machine learning methods by integrating AutoML with Data Pipeline. We will discuss how to leverage AutoML technology to enhance the intelligence of Data Pipeline, thereby achieving better results in machine learning tasks. By delving into the automation and optimization of Data flows, we uncover key strategies for constructing efficient data pipelines that can adapt to the ever-changing data landscape. This not only accelerates the modeling process but also provides innovative solutions to complex problems, enabling more significant outcomes in increasingly intricate data domains.

## I. INTRODUCTION

The use of Machine Learning techniques and methods to solve practical problems has been successfully applied to many fields, and we often see examples of personalized recommendation systems, financial anti-fraud, natural language processing and machine translation, pattern recognition, intelligent control and so on. typical machine learning process usually includes source data ETL, data preprocessing, index extraction, model training, cross-validation, and new data prediction, among others. In today's real-world data work, the data we need to deal with is often diverse[1]. For example, imagine this scenario: if we need to do some analysis of a product, the source of data may be from social media user reviews, click rates, or transaction data obtained from sales channels, or historical data, or product information captured from product websites. In the face of so many different data sources, the data you have to deal with may include CSV files, may also have JSON files, Excel and other forms, may be pictures and text, may also be stored in the database table, and may be from the website, APP real-time data. This paper focuses on data pipeline training: Integrated AutoML takes optimizing the data flow of machine learning model as its core content, and analyzes the integrated optimization model and experimental process of data pipeline.

## II. OVERVIEW OF RELATED CONCEPTS

### A. Data pipeline definition and function

A data pipeline is a system used to automate the processing and transmission of data. It can extract the data from the source system, through cleaning, conversion, loading and other steps, and then transfer to the target system. The main function of data pipeline is to improve the efficiency of data processing, ensure the accuracy of data, and ensure the security of data. In traditional data processing, manual operations take up most of the time. The data pipeline through the automatic way, can greatly improve the efficiency of data processing. For example, a data pipeline can automate data processing at regular intervals by setting up scheduled tasks. In this way, it can not only reduce the time and effort of manual operation, but also avoid data errors caused by human factors.

Secondly, the data pipeline has high security and accuracy. In the process of data processing, due to various reasons, such as the inaccuracy of the data source and errors in the data transmission process, the data may be inaccurate. The data pipeline can verify and clean the data in real time by setting the data verification rules, so as to ensure the accuracy of the data.

---

[1] * Corresponding author: [Jiang Wu]. Email: [jiangwu@usc.edu]

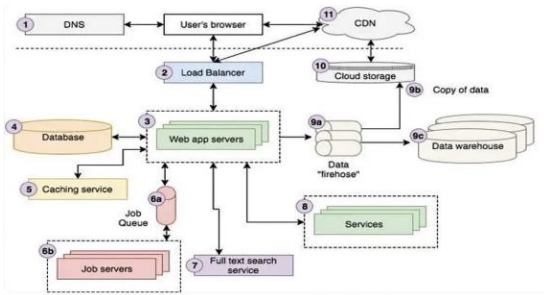

Figure 1. Precise data pipeline optimization steps

Overall, the data pipeline plays an important role in the modern enterprise. It can not only improve the efficiency of data processing, ensure the accuracy of data, but also ensure the security of data. Therefore, for any enterprise, it should pay attention to and invest in the construction and use of data pipelines.

### B. Data Pipeline and machine learning

There are many applications of Data Pipeline, and I will mainly introduce its application in machine learning. Although the application in the field of machine learning is only a small application of Data Pipeline, it is very successful[2]. For machine learning, the main task of Data Pipeline is to enable machines to analyze existing data, so that machines can make reasonable judgments on new data.

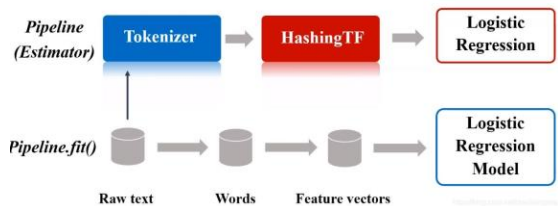

Figure 2. Pipeline model

In machine learning, there are usually a series of algorithms that process and learn from data. For example, a simple data text processing process can be divided into the following stages: (1) dividing text into words; (2) converting words into numerical feature vectors[3]; (3) Obtaining a prediction model through feature vectors and label learning. MLlib defines the above workflow as a Pipeline, which contains a series of Pipelinestages (Transformers and Estimators) running in a specific order.

### C. Pipeline Analysis of the execution process

The Pipeline object accepts a list of binary tuples, with the first element in each binary tuple being an arbitrary identifier string. The second element in the binary tuple is the transformer or estimator that scikit-learn ADAPTS to the individual elements in the access Pipeline object.

Code: Pipeline([('sc', StandardScaler()), ('pca', PCA(n_components=2)), ('clf', Logistic Regression(random_state=1))])

The middle of the Pipeline consists of a scikit-learn adaptor transformer, and the final step is an estimator[4]. For example, in the above code, StandardScaler and PCA transformer form intermediate steps, and LogisticRegression acts as the final estimator.

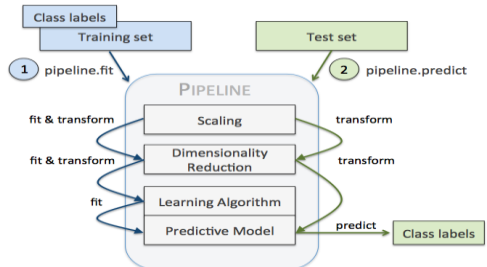

Figure 3. Diagram of data pipeline steps

When we execute pipe_lr.fit(X_train, y_train), the fit and transform methods are first executed by StandardScaler on the training set, and the transformed data is passed to the next step of the Pipeline object. Also known as PCA(). Like StandardScaler, PCA performs the fit and transform methods and ultimately passes the transformed data to LosigsticRegression.

### D. AutoML Introduction

Automatic Machine Learning (AutoML) is the process of automating end-to-end processes that apply machine learning to real-world problems[5]. The traditional machine learning model can be roughly divided into the following four parts: data acquisition, data preprocessing, optimization, application; The data preprocessing and model optimization part often requires data scientists with professional knowledge to complete, they build a bridge between data and computation. However, even data scientists need to spend a lot of effort to choose algorithms and models. The success of machine learning in various applications has led to a growing demand for machine learning practitioners, so we want to achieve true machine learning, so that as many jobs as possible can also be automated, further lowering the barrier to machine learning, so that people without expertise in the field can also use machine learning to complete the relevant work[6].

Figure 4. AutoML machine learning applications

Starting from the traditional machine learning model, AutoML realizes automation from three aspects: feature engineering, model construction and hyperparametric optimization. And also put forward the end-to-end solution. This column, the implementation of the idea of AutoML[7], to minimize the threshold, a brief introduction to the principle, focusing on the use of AutoML open source tools.

*E. Model structure optimization*

AutoML for Model Compression(AMC)[3] is a framework that uses reinforcement learning to automatically search and improve the quality of model pruning algorithms. The complete process is as follows:

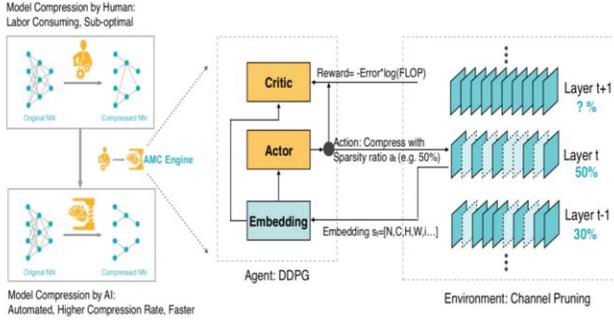

Figure 5.  Model optimization quality framework

In the general pruning algorithm, we usually follow some basic strategies: for example, cut fewer parameters in the first layer with fewer parameters to extract low-level features, and cut more parameters in the FC layer with higher redundancy. However, since the layers in a deep neural network are not isolated, these rule-based pruning strategies are not optimal and cannot be migrated from one model to another.

In this context, the AMC method uses reinforcement learning to automatically search for and improve the quality of model compression. In this framework, each layer is compressed independently and propagates to the next layer after the previous layer has been compressed. Layer t accepts the input feature $s\_t$ of this layer and outputs the sparse ratio $a\_t$. Use the accuracy of the validation set as an evaluation.

## III. METHODOLOGY AND RESULTS

Automated machine learning technology is a very important basic research, and it is also a hot direction in the optimization of deep learning model. In the relevant research of data pipeline, data enhancement is very important and an essential medicine for deep learning[8]. Therefore, if the model automatically learns data enhancement strategies for specific tasks, it will be more intelligent in theory. This is data enhancement technology based on AutoML.

Based on this, this paper is based on Pycaret automatic machine learning library on the basis of data pipeline, and demonstrates its various API usage and step flow.

*A. AUTOML Classification & Regression*

In the process of data optimization, the Pycaret library needs to be utilized, and there is no need to pay too much attention to the complex data preprocessing process during screening. IIntelligent screening, feature selection, model building, and structural analysis can be conducted through Pycaret, the comprehensive automation tool for AutoML, all at once.

The core steps are:

Setup → Compare Models → Analyze Model → Prediction → Save Model

Setup:

loading sample dataset from pycaret dataset module

from pycaret.datasets import get_data

data = get_data('diabetes')

Figure 6.  step function output result

For the setup function, we need to input at least two things: the data itself, and the Target tag. If the other parameters remain default, the configuration table can be generated, as shown in the figure above[9]. The whole configuration process involves the division of training test, pre-processing process, missing value process, cross verification part and so on. If we do not select, it is automatically adjusted according to the default. In addition to these two specific parameters, the experiment also makes the parameter analysis of the whole database function and obtains the corresponding model.

*B. Contrast model*

Setup: # compare baseline models

best = compare_models()

Figure 7.  Compare the model result parameters

After running the comparison model function, the algorithm returns the best model and the result of the comparison. So many algorithms are defined in the classification. In addition, the index is classified and classified, and two parameters are extracted as the main parameters, such as sort parameter. The default classification is accuracy rate and n_select parameter indicates how many models are selected. The rest includes cross-verification methods, and include exclude parameter indicates which models are used or excluded.

*C. Model introduction*

The analysis model is mainly plot_model function, among which the plot_model function supports a variety of drawing and the analysis result is more suitable for this experiment[10]. The function is relatively simple in the main bit and has two main parameters, the first is the model and the second is the plot parameter. The following is a graphical illustration of the results of several parameters:

Setup: plot_model(best,plot='gain')

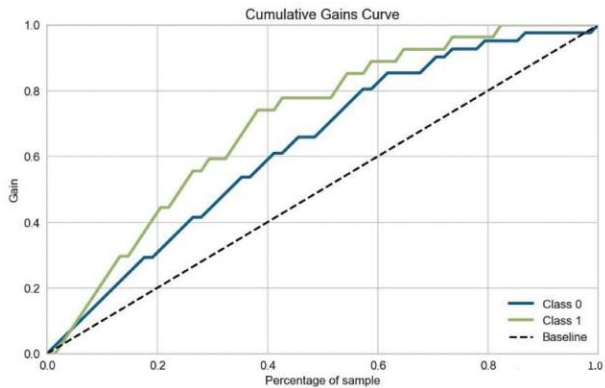

Figure 8. Parameter optimization results 1

Setup2: plot_model(best[0],plot='learning')

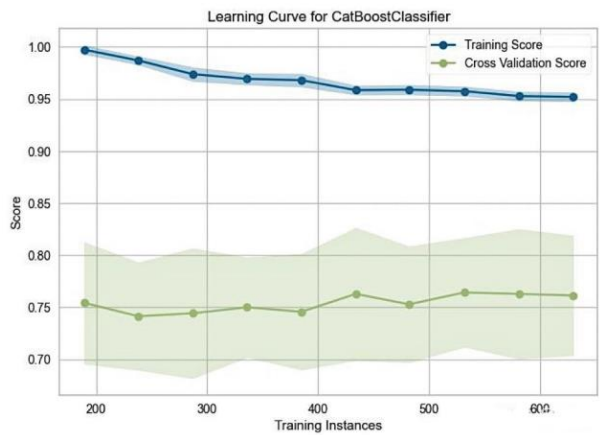

Figure 9. Parameter optimization results 2

Setup:plot_model(best[0],plot='feature')

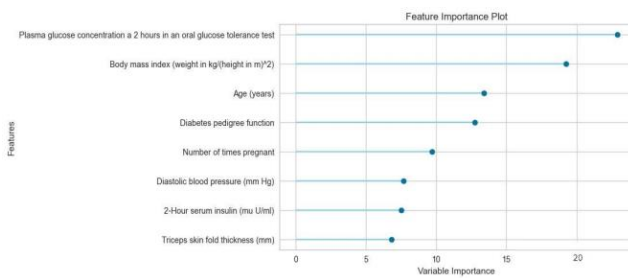

Figure 10. Parameter optimization results 3

After comparing the models, the core model needs to be saved. Among all parameters of save_model for the specific model, the first three parameters are mainly used. If the third parameter is set to True, only the trained model will be saved instead of the whole machine learning flow. Default is recommended.

*D. Set up individual models and hyperparameter tuning*

Suppose that to parameter adjust and optimize the selected model, it is necessary to create a separate model for the pycaret database, which requires the creat_model() function.

Setup:# train logistic regression with default fold=10

lr = create_model('lr')

| Fold | Accuracy | AUC | Recall | Prec. | F1 | Kappa | MCC |
|---|---|---|---|---|---|---|---|
| 0 | 0.8148 | 0.9023 | 0.5789 | 0.8462 | 0.6875 | 0.5624 | 0.5828 |
| 1 | 0.8333 | 0.7970 | 0.6316 | 0.8571 | 0.7273 | 0.6112 | 0.6260 |
| 2 | 0.8519 | 0.9383 | 0.6316 | 0.9231 | 0.7500 | 0.6499 | 0.6736 |
| 3 | 0.7222 | 0.7759 | 0.4211 | 0.6667 | 0.5161 | 0.3350 | 0.3524 |
| 4 | 0.8333 | 0.9083 | 0.5789 | 0.9167 | 0.7097 | 0.6010 | 0.6322 |
| 5 | 0.6852 | 0.6737 | 0.4211 | 0.5714 | 0.4848 | 0.2656 | 0.2720 |
| 6 | 0.7222 | 0.7820 | 0.4737 | 0.6429 | 0.5455 | 0.3520 | 0.3605 |
| 7 | 0.7547 | 0.8460 | 0.3333 | 0.8571 | 0.4800 | 0.3579 | 0.4263 |
| 8 | 0.7358 | 0.6952 | 0.4444 | 0.6667 | 0.5333 | 0.3592 | 0.3736 |
| 9 | 0.7358 | 0.7492 | 0.4444 | 0.6667 | 0.5333 | 0.3592 | 0.3736 |
| Mean | 0.7689 | 0.8068 | 0.4959 | 0.7614 | 0.5968 | 0.4453 | 0.4673 |
| Std | 0.0557 | 0.0857 | 0.0970 | 0.1236 | 0.1024 | 0.1353 | 0.1379 |

Figure 11. Model cross-validation results

*E. Experimental summary:*

This experiment is based on Pycaret automatic machine learning library and emphasizes the importance of applying automated machine learning in the data pipeline. By loading the sample dataset 'diabetes' and using Pycaret's configuration and comparison functions, we demonstrate the critical role of automatic machine learning in the data optimization process. From model comparison and visualization to preservation and deployment, Pycaret provides an integrated solution that enables data scientists to configure, compare and optimize models more efficiently.

Further, we highlight the flexibility of automated machine learning techniques in individual model tuning and hyperparameter optimization. By creating individual model instances, we are able to personalize specific models to further improve model performance. Overall, the integrated advantages of automatic machine learning make the automation of the data pipeline more complete and provide strong support for processing complex machine learning tasks. This experiment provides deep insights into the integration of automated machine learning into the data pipeline and is expected to open up new possibilities for future developments in the field of data science and machine learning.

IV. DISCUSSION AND CONCLUSION

*A. AUTOML Machine Learning automation benefits*

With AUTOML's PyCaret-based machine learning automation, data scientists can achieve more efficient model development in the data pipeline. Pycaret simplifies the entire machine learning process, from data preprocessing to model comparison, providing users with simple and powerful tools[12]. The built-in automation significantly improves model performance through steps such as algorithm selection, feature engineering, and hyperparameter tuning. This automation can shorten the model development cycle, reduce human intervention, and make machine learning more efficient.

The synergy of AutoML technology with Data Pipeline emerges as a key focal point, illustrating how this integration elevates the intelligence of the Data Pipeline. This strategic fusion empowers organizations to achieve superior results in various machine learning tasks. By delving into the intricate interplay between AutoML and Data Pipeline, this research elucidates nuanced strategies that are instrumental in crafting adaptive and efficient data pipelines. These strategies not only expedite the modeling process but also position organizations to devise innovative solutions to the multifaceted challenges inherent in the ever-evolving data landscape.

*B. Data pipeline data optimization benefits*

Data optimization will play a key role in the future of data analytics. With AUTOML and Pycaret, automated optimization of data pipelines not only improves model performance, but also lowers the technical barrier for data scientists[15]. In the future, we can expect more automated tools that can adapt to data pipelines in different fields and industries for a wider range of applications. In addition, the optimization of the data pipeline will become an important part of enabling intelligent decision making and innovation, providing more reliable, interpretable and repeatable analytical results for the business.

*C. The outlook for data optimization in the data pipeline*

As the volume of data increases and business needs continue to evolve, data pipeline data optimization will become a trend in the field of data analysis[16-17]. The continuous evolution and improvement of automated machine learning tools will drive efficiency and quality improvements in the data pipeline. In the future, we expect to see more data pipeline optimization solutions that focus on interpretability, continuous optimization, and adapting to the ever-changing data environment. This will enable enterprises to make better use of data resources and adapt more quickly to market changes, thus driving further development in the field of data analytics.

ACKNOWLEDGMENT

At the end of the article, I want to tell Liu, Bo et al., "Integration and Performance Analysis of Artificial Intelligence and Computer Vision Based on Deep Learning Algorithms" (arXiv:2312.12872) for the excellent research presented. Their work not only provides me with valuable reference and inspiration for this article, but also provides deep insights for research in the fields of data pipeline automation, machine learning, and data optimization. The contribution of this paper not only enriches the academic community's understanding of integrated AI and computer vision, but also provides a solid foundation for my discussion in the article.